%% file: main.tex
\documentclass[letterpaper,journal]{IEEEtran}
\usepackage{amsmath,amssymb}
\usepackage{algorithmic}
\usepackage{algorithm}
\usepackage{array}
\usepackage{booktabs}
\usepackage{multirow}
\usepackage[caption=false,font=normalsize,labelfont=sf,textfont=sf]{subfig}
\usepackage{textcomp}
\usepackage{url}
\usepackage{verbatim}
\usepackage{graphicx}
\usepackage{xcolor}
\usepackage{cite}
\newcommand{\gaincell}[2]{#1\,{\scriptsize\textcolor{black!55}{(+#2)}}}

\makeatletter
\newcommand{\closerband}{\noalign{{\color{gray!12}%
\hrule height 0pt depth \dimexpr\ht\@arstrutbox+\dp\@arstrutbox\relax width \textwidth}%
\vskip-\dimexpr\ht\@arstrutbox+\dp\@arstrutbox\relax}}
\makeatother

\begin{document}

\title{CLOSER: Evidence-Aware Active Visual Verification for Embodied Object-Goal Navigation}

\author{Pengying~Wu,
        Mingye~Jiang,
        Yunze~Hu,
        Zhejia~Wen,
        Kangjie~Zhou,
        Zongru~Yang,
        Zhe~Peng,
        and~Chang~Liu%
\thanks{Pengying Wu, Mingye Jiang, Yunze Hu, Zhejia Wen, Kangjie Zhou, Zongru Yang, Zhe Peng, and Chang Liu are with the School of Advanced Manufacturing and Robotics, Peking University, Beijing 100871, China (e-mail: changliucoe@pku.edu.cn).}%
\thanks{(Corresponding author: Chang Liu.)}}

\markboth{}%
{Wu \MakeLowercase{\textit{et al.}}: CLOSER: Evidence-Aware Active Visual Verification for Embodied Object-Goal Navigation}

\maketitle

\begin{abstract}
Object-goal navigation (ObjectNav) requires an embodied agent to find an instance of a specified category in an unseen environment and stop near it. The agent typically relies on an object detector but must judge online from a limited egocentric video stream, sometimes a single frame, whether a detected candidate is the target. Detectors often confidently mistake similar objects, such as a sofa for a bed, and stopping at a false positive fails the episode. Prevailing systems also approach candidates greedily, leaving them occluded or partially observed. We present CLOSER, an evidence-aware active verification framework that treats stopping as closed-loop evidence acquisition, coupling vision--language model (VLM) verification with active perception. CLOSER fine-tunes an open-source VLM to label each view of a candidate tracked across frames as target, non-target, or uncertain, selects complementary views with an embedding-based diversity gate, and fuses their per-view probabilities with a lightweight network. If the fused judgment remains uncertain, a diffusion policy drives the agent toward the candidate using only the tracked mask and depth until the decision becomes decisive. CLOSER runs locally and plugs into existing ObjectNav systems without retraining them. On Gibson and Habitat-Matterport 3D, it raises average success rate by 3.6 points and success weighted by path length by 2.0 points, showing that verifying evidence before committing improves embodied navigation.
\end{abstract}

\begin{IEEEkeywords}
Object-goal navigation, embodied AI, spatial reasoning, scene understanding, vision for robotics, object detection and recognition.
\end{IEEEkeywords}

\input{sections/introduction}

\input{sections/related_work}

\input{sections/methodology}

\input{sections/experiments}

\input{sections/conclusion}

\section*{Acknowledgment}
The authors declare that they have no known competing financial interests or personal relationships that could have appeared to influence the work reported in this paper.

\bibliographystyle{IEEEtran}
\bibliography{main}

\end{document}

%% file: sections/introduction.tex
\section{Introduction}

\begin{figure}[t]
\centering
\includegraphics[width=\columnwidth]{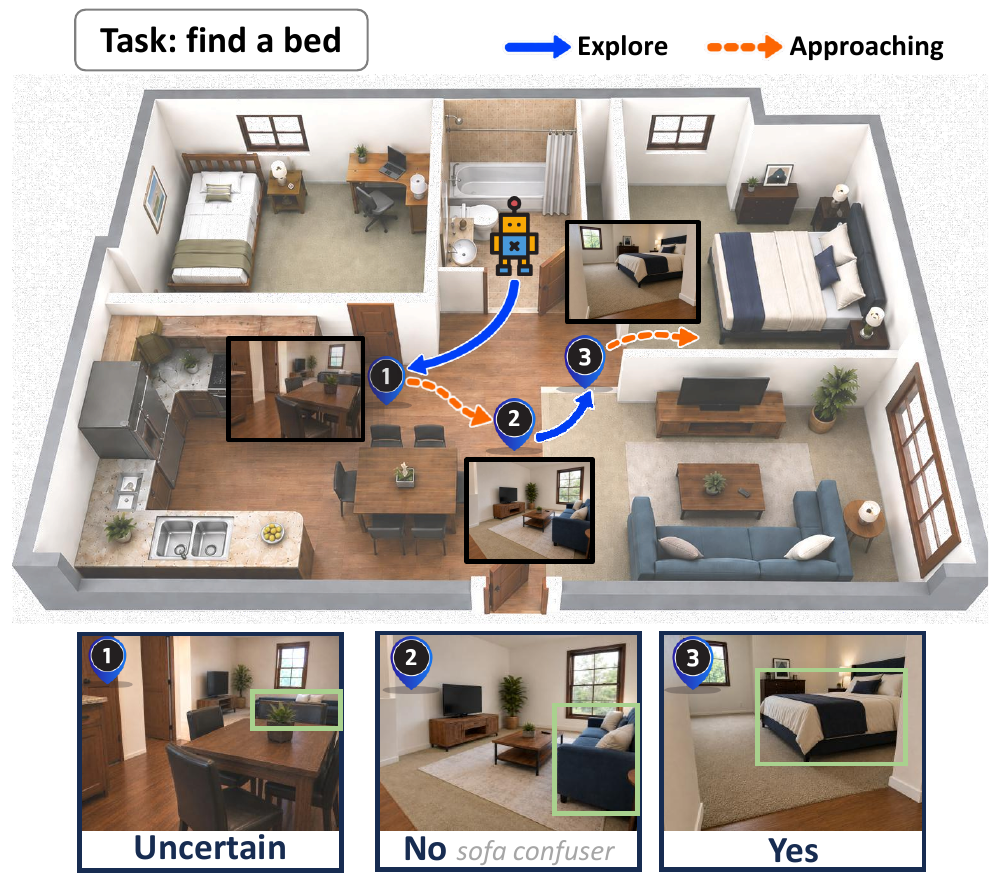}
\caption{\textbf{With CLOSER, the agent approaches an uncertain candidate and decides from closer views.}
Given the goal \emph{bed}, the view at (1) shows a distant, partially occluded candidate and does not suffice for a decision.
CLOSER therefore labels the candidate \texttt{uncertain}, and the agent approaches it to gather additional views (orange, dotted).
At (2), the closer view reveals a sofa confuser; CLOSER labels it \texttt{no}, and the agent resumes exploration (blue, solid).
At (3), the view of the next candidate suffices; CLOSER labels it \texttt{yes}, and the agent moves to the bed and stops near it.
The bottom panels show the agent's view and CLOSER's label at each numbered location.}
\label{fig:teaser}
\end{figure}

\IEEEPARstart{O}{bject-goal} navigation (ObjectNav) is a fundamental task in embodied AI~\cite{11098567} in which an agent searches an unseen environment for an object of a specified target category and stops near it~\cite{chaplot2020semexp,fukushima2023omt}. To find the target, the agent must explore the environment efficiently and infer where the target is likely to be located. When an object resembling the target comes into view, the agent must further judge whether it is the target. ObjectNav therefore demands both efficient exploration and reliable perception from its online egocentric video stream, and remains a challenging problem despite extensive study~\cite{sun2024objectnav_survey}.

To meet these demands, prevailing modular systems adopt an exploration strategy to search promising regions and an off-the-shelf object detector to recognize the observed objects~\cite{chaplot2020semexp,zhou2023esc,yokoyama2024vlfm}. Whenever a candidate for the target category is detected, the agent must decide from limited local observations whether the candidate is the target and hence whether to stop; in the worst case, only a single partial view is available. Since the episode ends once the agent stops, a wrong stop cannot be corrected, and this stopping decision has been identified as a major bottleneck of modular systems~\cite{song2023dita,ratenav2025}. Although modern open-vocabulary detectors such as YOLO-World and Grounded-SAM~\cite{cheng2024yolo,ren2024grounded} achieve high recall on common indoor categories, they frequently mistake semantically similar objects for the target. For example, a sofa observed from a distance may be detected as a bed (Fig.~\ref{fig:teaser}), and a stool partially occluded by a table may be detected as a chair. Such \emph{confusers} are often detected with tight bounding boxes and high confidence, so raising the confidence threshold removes true targets before it removes these errors. Therefore, each detected candidate needs to be further verified before the agent commits to a stop.

To this end, recent systems introduce a closed-source vision--language model (VLM) such as GPT-4o-mini as a verifier that confirms or rejects each detected candidate~\cite{zhu2025strive}. Such a model provides open-world visual reasoning that the detector lacks, and can reject confusers even when the detection confidence is high. However, the verifier is accessible only through a remote API, which incurs per-query cost and latency and leaves its weights unavailable for local adaptation. A smaller open-source VLM avoids these constraints, yet its verification accuracy drops markedly on hard cases such as distant or partially occluded confusers. Moreover, existing verifiers are binary and must decide even when the current view supports neither confirmation nor rejection. Such a forced decision may stop the agent at a non-target or reject a candidate that a closer view could confirm. Recent work has proposed ternary uncertainty scores and explicit \textsc{not-found} outcomes~\cite{qasknav2026,vlnnf2026}, but neither mechanism moves the agent to acquire more views of the candidate for verification.

In this paper, we propose \textbf{CLOSER} (Closed-Loop Observation Sensing for Embodied veRification), an evidence-aware active verification framework in which the agent decides whether to stop by reasoning over the visual evidence on a candidate and acting to improve it. Our key insight is that a reliable stopping decision depends not only on what the candidate appears to be but also on whether the available views suffice to identify it, and this sufficiency depends on the spatial relationship between the agent and the candidate, including its distance, viewing angle, and degree of occlusion. The agent therefore commits to a decision only when the evidence suffices and otherwise approaches the candidate to acquire more informative views (Fig.~\ref{fig:teaser}). Realizing this closed loop poses two challenges. First, the agent must learn \emph{when} its evidence is insufficient, which confidence cannot reveal since distant or occluded views often yield confident predictions. Second, it must learn \emph{how} to reach a more informative view without a complete map of the unseen environment.

To learn \emph{when} the evidence is insufficient, CLOSER fine-tunes an open-source VLM with LoRA~\cite{hu2022lora} to assess both the identity of the candidate and the sufficiency of each view. During navigation, CLOSER tracks the candidate across egocentric frames with ByteTrack~\cite{zhang2022bytetrack} and caches complementary views of it, and a lightweight MLP fuses the per-view assessments of the VLM to decide whether the evidence suffices. To learn \emph{how} to reach a more informative view, CLOSER imitates an expert that plans maximum-clearance paths along the Voronoi medial axis of the complete map and smooths them into B\'ezier trajectories, so that walls are less likely to occlude the candidate along the way. To eliminate this dependence on the map, we distill the expert trajectories into a Diffusion Transformer policy~\cite{chi2023diffusion_policy,peebles2023dit} that conditions only on the tracked candidate mask and depth, which encode the distance and visibility of the candidate relative to the agent. As the agent follows this policy, verification and active perception form a closed loop, in which every view cached along the trajectory refines the assessment and the approach phase ends as soon as the evidence supports a decision. CLOSER is invoked only when a candidate for the target category is detected and leaves the exploration and mapping components of existing ObjectNav systems unchanged. It can therefore be inserted into these systems as a plug-in verification layer without retraining them.

The main contributions of this paper are summarized as follows:
\begin{itemize}
\item We propose CLOSER, a plug-in layer for existing ObjectNav systems that formulates the stopping decision as an active verification problem. Closer judges not only whether a candidate is the target but also whether its views suffice for that judgment, and it drives the agent to acquire more views by approaching the candidate when the current views are insufficient.
\item Since a confident prediction on a distant or occluded view can be wrong, we design the verifier in Closer to assess the sufficiency of each view in addition to the identity of the candidate. We fine-tune an open-source VLM for this assessment under a closed-source VLM teacher and train a lightweight MLP that fuses the assessments of complementary views into a single decision.
\item We train a map-free approach policy by distilling an expert into a Diffusion Transformer that generates a trajectory toward the candidate from its tracked mask and depth. To keep the candidate in view throughout the approach phase, the expert plans each path along the Voronoi medial axis of the free space for maximum clearance from walls and smooths the path into a B\'ezier trajectory.
\item Simulation experiments confirm that CLOSER improves ObjectNav performance. Across diverse ObjectNav systems on Gibson, HM3D v1, and HM3D v2, CLOSER improves success rate (SR), with an average gain of 3.6 points and a maximum gain of 4.9 points. CLOSER also improves success weighted by path length (SPL) by 2.0 points on average.
\end{itemize}

%% file: sections/related_work.tex
\section{Related Work}
\label{sec:related_work}

\subsection{Object-Goal Navigation and Target Verification}
\label{subsec:rw_objectnav}

Existing ObjectNav methods range from end-to-end reinforcement learning, which learns continuous environment representations with multi-step look-ahead planning~\cite{chen2024think} or agent-centric relation graphs that encode object layout and target distance~\cite{hu2024acrg}, to modular pipelines that combine semantic mapping with goal selection~\cite{chaplot2020semexp,fukushima2023omt} and zero-shot variants~\cite{sun2024objectnav_survey}, supported by large-scale simulators~\cite{yadav2023hm3dsem}. Recent methods use vision--language models to score exploration frontiers~\cite{yokoyama2024vlfm} or large language models to prioritize search regions~\cite{zhu2025chatnav,cao2025cognav}; E$^2$BA~\cite{shi2025e2ba} further fuses LLM commonsense scores with geometric costs through particle filtering and uses a cascaded discriminator to decide when to backtrack. These advances concentrate on where to explore, and a separate mechanism is still required to verify whether a detected candidate is the target. For the stopping decision, DITA~\cite{song2023dita} decouples depth-based termination from the navigation policy, RATE-Nav~\cite{ratenav2025} models spatial regions to trigger stopping, and ACRG~\cite{hu2024acrg} encodes the depth-derived agent--target distance into the policy. Classical pipelines verify candidates with detector confidence thresholds~\cite{chaplot2020semexp}, which fail under partial observability~\cite{rememNav2026}. Memory~\cite{fukushima2023omt,rememNav2026} and blacklist mechanisms~\cite{yin2025unigoal} improve decision consistency, SPENav~\cite{yuan2026spenav} filters task-irrelevant objects with room-guided semantic priors, and recent systems query a closed-source VLM to verify each candidate~\cite{zhu2025strive}. For image goals, RSRNav~\cite{qin2026rsrnav} reasons about the spatial relationship between the goal and the current observation through direction-aware correlation and remains robust when the goal viewpoint differs from the test viewpoint. Beyond binary decisions, QAsk-Nav~\cite{qasknav2026} and VLN-NF~\cite{vlnnf2026} introduce ternary uncertainty scores and an explicit \textsc{not-found} outcome, while IEVE~\cite{lei2024ieve} and DRIVE-Nav~\cite{gao2026drivenav} verify potential targets by keypoint matching and cross-frame VLM inspection. However, these systems either depend on closed-source VLM APIs or do not collect further observations when the evidence on a candidate is insufficient. In CLOSER, verification runs on local hardware, and the agent approaches a candidate with insufficient evidence to acquire additional views.

\subsection{Vision--Language Model Adaptation and Chain-of-Thought Reasoning}
\label{subsec:rw_vlm}

Large VLMs can be adapted by training only a small fraction of their parameters, either a lightweight query module with the image encoder and language model frozen~\cite{dai2023instructblip} or low-rank updates such as LoRA~\cite{hu2022lora}, and EverWalker~\cite{li2026everwalker} generates LoRA parameters on the fly from observation prototypes for lifelong navigation. Chain-of-thought supervision provides training signal beyond the final label: CL-CoTNav~\cite{cai2025cl} fine-tunes a VLM on hierarchical chain-of-thought annotations so that it reasons about ObjectNav decisions, and distilling rationales from closed-source teachers into open models is now standard practice~\cite{zhang2024vlmkd,li2023scotd}. Distillation is likewise used to compress navigation models: Chen et al.~\cite{chen2026qad} distill both where to look and how to move from a teacher into a student with 93\% fewer parameters. However, CURE~\cite{chen2024cure} shows that VLM rationales can be inconsistent with the visual evidence even when the final answer is correct, and program-based distillation~\cite{hu2024vpd} mitigates this inconsistency through executable verification. CLOSER adapts Qwen2.5-VL-7B with LoRA and keeps a base-model rationale only when its label is correct and a rule-based validator accepts it, so an external teacher annotates only the remaining samples (Sec.~\ref{subsec:cot_sft}). Rendering novel viewpoints with 3D Gaussian Splatting also improves the viewpoint robustness of VLN agents~\cite{zhong2026spatial}; likewise, CLOSER renders confusers as hard negatives and refines the perceiver on its residual errors (Sec.~\ref{subsec:render_augmentation}).

\subsection{Active Perception and Uncertainty-Driven Exploration}
\label{subsec:rw_active}

Active perception selects sensor placements to maximize information gain. Next-best-view (NBV) methods optimize hand-crafted metrics~\cite{breyer2022nbv} or predict the downstream value of prospective viewpoints~\cite{gennbv2024,jin2023neunbv}. Uncertainty-aware methods complement view selection through conformal prediction~\cite{ren2023knowno} and uncertainty-driven intrinsic motivation~\cite{seal2021}, and recent work applies them to open-vocabulary object search~\cite{uncertainty_informed2025}. Diffusion models have also become expressive policy representations that capture multimodal action distributions~\cite{chi2023diffusion_policy} and generate navigation trajectories~\cite{nomad2024}. The approach policy of CLOSER connects these two lines without an explicit information-gain objective (Sec.~\ref{subsec:active_sensing}). A Diffusion Transformer~\cite{peebles2023dit} distilled from privileged expert trajectories takes only the candidate mask and depth as input, and the verification stack decides from the newly acquired views whether the evidence suffices.

%% file: sections/methodology.tex
\section{Methodology}
\label{sec:methodology}

\subsection{Problem Formulation}
\label{subsec:problem_formulation}

We consider the standard ObjectNav setting, in which an agent placed in a previously unseen indoor environment must find an instance of a given target category $c$. At each step, the agent receives an egocentric RGB-D observation together with its pose relative to the start of the episode. It then executes either a discrete movement action or the stop action, which ends the episode. The episode succeeds if the agent calls the stop action within $1$\,m of an instance of the target category under a budget of $500$ steps.

A \emph{candidate event} begins with the detection of a candidate for the target category and ends when the agent decides to stop at the candidate or to reject it and resume exploration. The agent may defer this decision to gather further views of the candidate. The event ends without a decision if the agent loses track of the candidate or the episode terminates first. Since deferral consumes steps from the episode budget, the goal in each event is to reach a correct decision with as few deferral steps as possible.

\begin{figure*}[t]
    \centering
    \includegraphics[width=\textwidth]{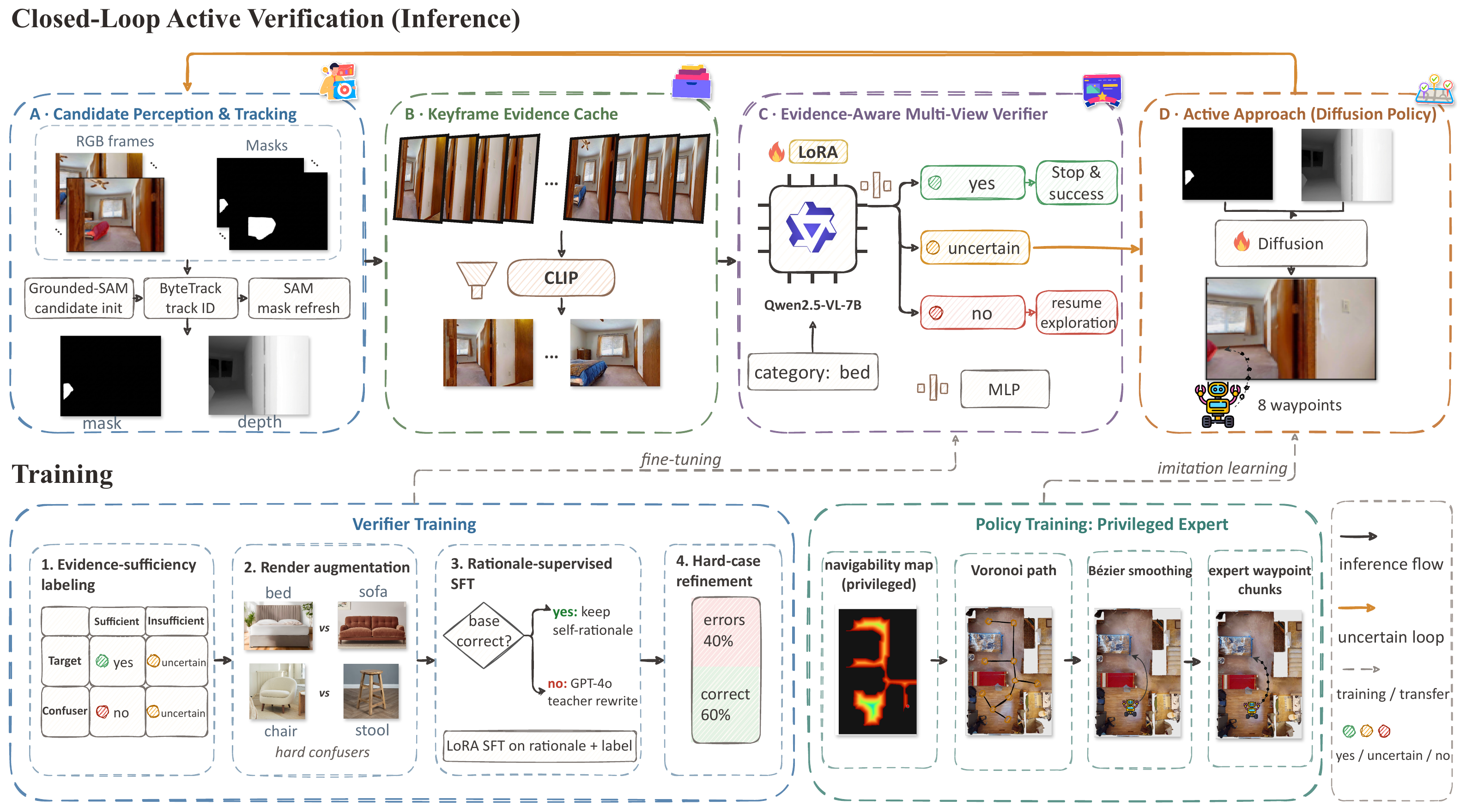}
    \caption{\textbf{Closed-loop inference and training of CLOSER.} (A) Grounded-SAM, ByteTrack, and SAM track the candidate and refresh its mask in each RGB-D frame. (B) The evidence cache retains views selected by a CLIP-based diversity gate. (C) Per-view tri-state probabilities from the single-view perceiver are fused by a lightweight MLP. (D) After an \texttt{uncertain} decision, a Diffusion Transformer generates waypoints from the mask and depth. Frames observed along the waypoints re-enter panels A and B (uncertain loop). The lower panels depict the perceiver training pipeline (\emph{Verifier Training}) and the construction of expert waypoint chunks (\emph{Policy Training}).}
    \label{fig:framework}
\end{figure*}

\subsection{Framework Overview}
\label{subsec:framework_overview}

CLOSER controls the agent only during a candidate event, so the host ObjectNav system keeps its exploration and mapping modules unchanged. As shown in Fig.~\ref{fig:framework}, the candidate is tracked across frames, and a CLIP-based diversity gate maintains an evidence cache of complementary views. The single-view perceiver is a fine-tuned VLM that predicts tri-state probabilities for each retained view (\texttt{yes} for a target, \texttt{no} for a non-target, and \texttt{uncertain} for insufficient evidence). A lightweight MLP fuses these probabilities into a tri-state decision (Sec.~\ref{subsec:evidence_verifier}). The perceiver and the MLP are trained offline on labels that encode both the semantic identity and the evidence sufficiency of each detection (Sec.~\ref{subsec:verifier_training}). If the fused decision is \texttt{uncertain}, the agent enters the approach phase, in which the approach policy generates a local trajectory toward the candidate from its tracked mask and depth. The policy is trained to imitate expert trajectories planned on a privileged navigability map (Sec.~\ref{subsec:active_sensing}). Views retained along the local trajectory update the fused decision, so the agent can end the approach phase before reaching the candidate (Sec.~\ref{subsec:closed_loop_inference}).

\subsection{Tri-State Verification Stack}
\label{subsec:evidence_verifier}

At each step $t$ of a candidate event, the verification stack outputs a fused decision $y_t\in\mathcal{Y}=\{\texttt{yes},\texttt{uncertain},\texttt{no}\}$ from the cached views. An \texttt{uncertain} decision indicates that these views support neither \texttt{yes} nor \texttt{no}, so targets and confusers alike can receive it. A binary verifier can also defer by withholding its decision whenever its confidence falls below a threshold. However, the threshold is selected on the validation split and is thus tied to the confidence distribution of that split. When this distribution shifts at deployment, confident predictions on insufficient views can exceed the threshold and become forced decisions, while no labels are available to re-tune the threshold. Moreover, a low confidence does not indicate whether additional views of the candidate would enable a decision. CLOSER therefore supervises \texttt{uncertain} as an explicit class. The training labels of this class are derived from view conditions that the agent can improve by moving, namely the image scale, truncation or occlusion, and viewing distance of the candidate (Sec.~\ref{subsubsec:labeling_function}).

\subsubsection{Candidate Tracking}
\label{subsubsec:candidate_tracking}

When Grounded-SAM~\cite{ren2024grounded} identifies a candidate for the target category in the RGB frame $I_t$, it returns a bounding box $\mathcal{B}_t$ and a segmentation mask $M_t$; $D_t$ denotes the aligned depth image. ByteTrack~\cite{zhang2022bytetrack} associates the candidate across subsequent frames, and SAM~\cite{kirillov2023segment} refreshes $M_t$ from the tracked box in each frame.

\subsubsection{Evidence Cache and Diversity Gate}
\label{subsubsec:keyframe_evidence}

Consecutive frames of a tracked candidate are often nearly identical, and admitting every frame would waste perceiver queries and bias the fused decision toward repeated views. The diversity gate therefore decides which frames enter the evidence cache of the tracked candidate. After frame $I_t$, the cache is
\begin{equation}
\label{eq:evidence_cache}
\mathcal{K}_t=\bigl\{(I^{(j)},\mathbf{v}^{(j)},\mathbf{p}^{(j)})\bigr\}_{j=1}^{m_t},\qquad m_t\leq K,
\end{equation}
where $m_t$ is the number of retained views and $K$ is the cache capacity. Entries are indexed in admission order, and entry $j$ stores the retained view $I^{(j)}$, its CLIP ViT-L/14~\cite{radford2021clip} feature $\mathbf{v}^{(j)}$, and its tri-state probabilities $\mathbf{p}^{(j)}$ from the perceiver (Sec.~\ref{subsubsec:single_view_perceiver}). The gate measures the novelty of the current frame by the minimum cosine distance between its CLIP feature $\mathbf{v}_t$ and the features in $\mathcal{K}_{t-1}$,
\begin{equation}
\label{eq:gate_similarity}
\rho_t = \min_{1\leq j\leq m_{t-1}} \bigl[1-\cos\bigl(\mathbf{v}_t,\mathbf{v}^{(j)}\bigr)\bigr],
\end{equation}
where $\cos(\cdot,\cdot)$ denotes cosine similarity. The frame is eligible only if $\rho_t$ exceeds the threshold $\delta_{\mathrm{view}}$ selected on the validation split. Each track starts with an empty cache, and its first frame is always eligible. If $m_{t-1}<K$, an eligible frame is appended as the newest entry $(I_t,\mathbf{v}_t,\mathbf{p}_t)$. If $m_{t-1}=K$, the gate keeps the size-$K$ subset of $\mathcal{V}_t=\{\mathbf{v}^{(1)},\ldots,\mathbf{v}^{(K)},\mathbf{v}_t\}$ with the largest diversity,
\begin{equation}
\label{eq:cache_update}
S_t^\star=\arg\max_{S\subseteq\mathcal{V}_t,\ |S|=K}\operatorname{Div}(S),
\end{equation}
where $\operatorname{Div}(S)$ is the mean cosine distance over all pairs of distinct features in $S$. If $\mathbf{v}_t\notin S_t^\star$, the frame is discarded without a perceiver query. Otherwise, the entry whose feature lies outside $S_t^\star$ is evicted, and the frame is appended as the newest entry.

\subsubsection{Single-View Perceiver}
\label{subsubsec:single_view_perceiver}

Each frame that enters the cache is evaluated exactly once by the perceiver $h_\theta$, a VLM that runs on local hardware and is fine-tuned as described in Sec.~\ref{subsec:cot_sft}. The perceiver receives the full frame $I_t$ and the target category $c$, so that the scale, occlusion, and context of the candidate remain visible. A constrained prompt restricts its output to the three labels. Their probabilities, renormalized over $\mathcal{Y}$, form $\mathbf{p}_t=h_\theta(I_t,c)\in\mathbb{R}^{3}$.

\subsubsection{Cross-View Fusion}
\label{subsubsec:fusion_mlp}

The fusion MLP $g_\omega$ receives the probability vectors of all cached views, which together retain how confident each view is and whether the views agree. Its input $\mathbf{P}_t$ stacks these vectors and fills the $K-m_t$ empty slots with zero rows,
\begin{equation}
\label{eq:probability_cache}
\mathbf{P}_t=\bigl[\,\mathbf{p}^{(1)},\ \ldots,\ \mathbf{p}^{(m_t)},\ \mathbf{0},\ \ldots,\ \mathbf{0}\,\bigr]^\top\in\mathbb{R}^{K\times3}.
\end{equation}
As every cached probability vector sums to one, a zero row marks an empty slot without a separate indicator. The MLP $g_\omega$ maps the flattened input $\operatorname{vec}(\mathbf{P}_t)\in\mathbb{R}^{3K}$ to logits $\mathbf{z}_t$ with one entry $z_{t,a}$ per label $a\in\mathcal{Y}$, and the fused decision is the label with the largest logit,
\begin{equation}
\label{eq:decision_function}
\mathbf{z}_t = g_\omega\bigl(\operatorname{vec}(\mathbf{P}_t)\bigr)\in\mathbb{R}^{3},\qquad
y_t = \arg\max_{a\in\mathcal{Y}} z_{t,a}.
\end{equation}

\subsection{Training the Verification Stack}
\label{subsec:verifier_training}

\subsubsection{Evidence-Sufficiency Labeling}
\label{subsubsec:labeling_function}

We assign training labels to Grounded-SAM detections, so that the training samples carry the same localization errors and imperfect masks as the candidates encountered at deployment. Each detection $(I,\mathcal{B},M,D)$ comprises the frame, box, mask, and depth image of Sec.~\ref{subsubsec:candidate_tracking} with the time index omitted. It is assessed along two axes. Semantic identity is derived from simulator annotations, whereas evidence sufficiency is computed from the detection alone.

Each detection is obtained by prompting Grounded-SAM with a target category $c$. Let $G_c$ be the pixels annotated as $c$ in the simulator, and let $|\cdot|$ count pixels. The semantic overlap is
\begin{equation}
\label{eq:target_ratio}
p_{\mathrm{target}}(M,G_c) = \frac{|M\cap G_c|}{|M|}.
\end{equation}
An overlap of at least $\eta_+$ indicates a target, an overlap below $\eta_-$ indicates a non-target, and intermediate values leave the identity ambiguous.

Evidence sufficiency is measured by three quantities that correspond to the view conditions of Sec.~\ref{subsec:evidence_verifier}: the mask-to-image ratio $r_{\mathrm{img}}=|M|/|I|$ for image scale, the box-fill ratio $r_{\mathrm{fill}}=|M|/|\mathcal{B}|$ for truncation or occlusion, and the mean mask depth $\bar d=\sum_{u\in M}D(u)/|M|$ for viewing distance. The sufficiency predicate requires all three quantities to satisfy their thresholds,
\begin{equation}
\label{eq:quality_predicate}
q(I,M,\mathcal{B},D)=
\mathbf{1}[r_{\mathrm{img}}\geq\alpha]\,
\mathbf{1}[r_{\mathrm{fill}}\geq\beta]\,
\mathbf{1}[\bar d\leq\gamma],
\end{equation}
where the indicator $\mathbf{1}[\cdot]$ equals $1$ if its condition holds and $0$ otherwise. The labeling function then assigns the label $\ell$ by checking sufficiency before identity,
\begin{equation}
\label{eq:label_assignment}
\ell =
\begin{cases}
\texttt{uncertain}, & q=0,\\
\texttt{yes}, & q=1 \wedge p_{\mathrm{target}}\geq\eta_+,\\
\texttt{no}, & q=1 \wedge p_{\mathrm{target}}<\eta_-,\\
\text{discard}, & \text{otherwise},
\end{cases}
\end{equation}
where $\eta_\pm$, $\alpha$, $\beta$, and $\gamma$ are the thresholds of the labeling function. Table~\ref{tab:two_axis} summarizes the four regimes of this assignment.

\begin{table}[t]
\centering
\caption{Evidence-aware label assignment.}
\label{tab:two_axis}
\begin{tabular}{@{}lcc@{}}
\toprule
 & Sufficient Evidence & Insufficient Evidence \\
\midrule
Target candidate & \texttt{yes} & \texttt{uncertain} \\
Non-target candidate & \texttt{no} & \texttt{uncertain} \\
\bottomrule
\end{tabular}
\end{table}

\subsubsection{Detector-Aligned Data and Render Augmentation}
\label{subsec:render_augmentation}

Natural training data come from collection trajectories that follow shortest paths through HM3D~\cite{yadav2023hm3dsem} training scenes in Habitat~\cite{savva2019habitat}. Grounded-SAM runs on sampled frames, and each detection is labeled by Eq.~(\ref{eq:label_assignment}). These data populate the four regimes of Table~\ref{tab:two_axis} unevenly. Most non-target detections are easy to reject, whereas sufficient views of confusers are scarce and insufficient views of confusers are rarer still. We therefore supplement these two regimes with rendered confusers, such as a sofa for the target \emph{bed} or a stool for the target \emph{chair} (Verifier Training panel of Fig.~\ref{fig:framework}). Each confuser is rendered with varied camera distance, azimuth, image-plane placement, and occlusion, so that it is observed in both sufficient and insufficient views. Rendered frames pass through the same detector and labeling function as natural frames. A rendered frame therefore yields a training sample only when Grounded-SAM identifies the confuser as a candidate for the target category, and the sample is labeled from the detected mask.

\subsubsection{Rationale-Supervised Adaptation and Refinement}
\label{subsec:cot_sft}

The perceiver is adapted from Qwen2.5-VL-7B-Instruct~\cite{bai2025qwen25vl} with LoRA~\cite{hu2022lora} on the resulting single-view corpus. Its training target is a short rationale followed by the tri-state label. The rationale names the category-relevant attributes visible in the frame and states whether the view suffices for a decision, so the supervision covers both axes of the labeling function. Since only the label is decoded at deployment, the rationale adds no inference latency.

To limit queries to a closed-source teacher, we first prompt the base Qwen model to generate a rationale $r_i^{\mathrm{base}}$ and a prediction $\hat\ell_i$ for each sample $i$ from its frame $I_i$ and category $c_i$, without access to its label $\ell_i$. A correct prediction can still be accompanied by a malformed or label-inconsistent rationale. A rule-based validator $V$ therefore checks each base rationale for format, length, label consistency, and prohibited patterns, and outputs $1$ only if all checks pass. The rationale used for training is
\begin{equation}
\label{eq:cot_triage}
r_i =
\begin{cases}
r_i^{\mathrm{base}}, & \hat\ell_i=\ell_i \wedge V(r_i^{\mathrm{base}},\ell_i)=1,\\
f_{\mathrm{ext}}(I_i,c_i,\ell_i), & \text{otherwise},
\end{cases}
\end{equation}
where $f_{\mathrm{ext}}$ queries GPT-4o to generate the rationale from the image, the category, and the label.

In the first round, the perceiver is trained with autoregressive cross-entropy over the rationale and label tokens. A hard-example refinement round then repeats the LoRA fine-tuning at a reduced learning rate. Its training set is drawn from the training split and combines every sample that the first-round perceiver misclassifies with correctly classified samples at a fixed ratio.

\subsubsection{Fusion MLP Training}
\label{subsubsec:fusion_mlp_training}

The fusion MLP is trained with the perceiver frozen. When a candidate stays tracked over several frames of a collection trajectory (Sec.~\ref{subsec:render_augmentation}), these frames pass through the diversity gate of Sec.~\ref{subsubsec:keyframe_evidence}, and the retained views form a multi-view training sequence. For sequence $i$, the perceiver probabilities of these views are stacked into $\mathbf{P}_i$ as in Eq.~(\ref{eq:probability_cache}). Each sequence is labeled by a best-evidence rule over the sufficiency values $q_j$ of its retained views, given by Eq.~(\ref{eq:quality_predicate}). A sequence without a sufficient view ($\max_j q_j=0$) is labeled \texttt{uncertain}. Otherwise, it is labeled \texttt{yes} or \texttt{no} by applying Eq.~(\ref{eq:label_assignment}) to its sufficient views. Sequences whose sufficient views yield ambiguous or conflicting identities are discarded. Since a single sufficient view makes the sequence label decisive, the MLP is trained to let that view outweigh the insufficient views cached with it. The training objective is three-class cross-entropy between $g_\omega(\operatorname{vec}(\mathbf{P}_i))$ and the sequence label.

\subsection{Active Approach Policy}
\label{subsec:active_sensing}

During the approach phase, the approach policy $\pi_\phi$ generates waypoints toward the candidate. Since shortest paths often pass close to walls and corners, where walls can block the line of sight to the candidate, the policy imitates expert trajectories smoothed from maximum-clearance paths.

\subsubsection{Privileged Expert Trajectory Construction}
\label{subsubsec:expert_trajectory}

Each expert demonstration starts from a simulated state in which a target instance is detected as a candidate whose view fails the sufficiency predicate of Eq.~(\ref{eq:quality_predicate}). The goal of the expert is obtained by back-projecting the candidate mask and depth to a target position and projecting this position onto the navigable plane. The expert has privileged access to the ground-truth navigability map (Policy Training panel of Fig.~\ref{fig:framework}), on which it plans a piecewise-linear path to the goal along the Voronoi medial axis of the free space. A B\'ezier curve fitted to this path yields a smooth trajectory, from which waypoints are sampled at a fixed spacing. Sliding windows of $H$ consecutive waypoints form the expert chunks. The chunk $\mathbf{A}_t^0\in\mathbb{R}^{H\times2}$ that starts at step $t$ stacks the planar offsets of its waypoints in the egocentric frame of the agent at step $t$.

\subsubsection{Mask-and-Depth-Conditioned Diffusion Transformer}
\label{subsubsec:diffusion_policy}

At deployment, the policy uses no map and observes only the tracked mask $M_t$ and the aligned depth image $D_t$, which encode the relative position, scale, and visibility of the candidate with respect to the agent. These inputs are concatenated channel-wise and processed by encoder $E_\psi$ into the condition $\mathbf{e}_t=E_\psi([M_t;D_t])$. Appearance reasoning is left to the perceiver, so the policy takes no RGB input. Omitting RGB is also intended to reduce the sensitivity of the policy to the appearance gap between simulation and reality. As the policy receives no category input, it approaches an \texttt{uncertain} confuser in the same way as an \texttt{uncertain} target, although all demonstrations end at targets.

Similar mask and depth observations in different room layouts can correspond to geometrically distinct expert chunks, and direct regression would average them. We therefore model the distribution of expert chunks conditioned on $\mathbf{e}_t$ with a Diffusion Transformer~\cite{peebles2023dit,chi2023diffusion_policy}. Each chunk $\mathbf{A}_t^0$ is paired with the condition $\mathbf{e}_t$ computed at its start step $t$. The transformer $\boldsymbol{\epsilon}_\phi$ predicts the Gaussian noise $\boldsymbol{\epsilon}\sim\mathcal{N}(\mathbf{0},\mathbf{I})$ added to $\mathbf{A}_t^0$ at diffusion timestep $k$ and is trained jointly with $E_\psi$ by minimizing
\begin{equation}
\label{eq:diffusion_loss}
\mathcal{L}_{\mathrm{DP}}(\phi,\psi) =
\mathbb{E}_{\mathbf{A}_t^0,k,\boldsymbol{\epsilon}}
\left[\left\|\boldsymbol{\epsilon}-\boldsymbol{\epsilon}_\phi(\mathbf{A}_t^k,k,\mathbf{e}_t)\right\|_F^2\right],
\end{equation}
where $\mathbf{A}_t^k=\sqrt{\bar\alpha_k}\,\mathbf{A}_t^0+\sqrt{1-\bar\alpha_k}\,\boldsymbol{\epsilon}$ is the noised chunk and $\bar\alpha_k$ is the cumulative noise schedule. At inference, reverse diffusion conditioned on the current mask and depth samples a chunk $\mathbf{A}_t\sim\pi_\phi(\cdot\mid M_t,D_t)$, and a low-level waypoint controller converts its waypoints into discrete movement actions.

\subsection{Closed-Loop Active Verification}
\label{subsec:closed_loop_inference}

Algorithm~\ref{alg:closed_loop} summarizes the inference procedure for one candidate event. Throughout the event, the perceiver evaluates each frame admitted by the diversity gate, and the fusion MLP updates the fused decision $y_t$ from the evidence cache (Sec.~\ref{subsec:evidence_verifier}). If $y_t$ is \texttt{yes} or \texttt{no}, the agent stops at the candidate or rejects it, respectively. Otherwise, the agent enters the approach phase and follows a chunk $\mathbf{A}_t$ sampled from the approach policy $\pi_\phi$ (Sec.~\ref{subsec:active_sensing}). Since verification continues during chunk execution, the agent cancels the remaining waypoints and acts on $y_t$ as soon as a newly retained view makes it decisive. When the agent completes a chunk while $y_t$ remains \texttt{uncertain}, the approach policy samples the next chunk from the current mask and depth. Once the mean mask depth $\bar d_t$ of $M_t$ falls below the close-range threshold $d_{\mathrm{close}}$, moving closer is unlikely to reveal substantially more detail. Therefore, if $y_t$ is still \texttt{uncertain} at this range, the agent applies a \emph{close-range fallback} that selects whichever of \texttt{yes} and \texttt{no} has the larger logit in $\mathbf{z}_t$.

\begin{algorithm}[!t]
\caption{Closed-Loop Active Verification}
\label{alg:closed_loop}
\begin{algorithmic}[1]
\REQUIRE target category $c$; RGB-D frames $(I_t,D_t)$; trained $h_\theta$, $g_\omega$, $\pi_\phi$; cache capacity $K$; view-diversity threshold $\delta_{\mathrm{view}}$; fallback depth $d_{\mathrm{close}}$
\STATE Grounded-SAM detects a candidate for $c$ in $I_t$; $\mathcal{K}_{t-1}\leftarrow\emptyset$
\WHILE{ByteTrack keeps tracking the candidate}
    \STATE Refresh $M_t$ with SAM from the tracked box; extract the CLIP feature $\mathbf{v}_t$ of $I_t$; $\mathcal{K}_t\leftarrow\mathcal{K}_{t-1}$
    \IF[Eqs.~(\ref{eq:gate_similarity}), (\ref{eq:cache_update})]{($m_{t-1}=0$ \OR $\rho_t>\delta_{\mathrm{view}}$) \AND ($m_{t-1}<K$ \OR $\mathbf{v}_t\in S_t^\star$)}
        \STATE If $m_{t-1}=K$, evict the entry outside $S_t^\star$ from $\mathcal{K}_t$
        \STATE $\mathbf{p}_t\leftarrow h_\theta(I_t,c)$; append $(I_t,\mathbf{v}_t,\mathbf{p}_t)$ to $\mathcal{K}_t$
    \ENDIF
    \STATE $\mathbf{z}_t\leftarrow g_\omega(\operatorname{vec}(\mathbf{P}_t))$; $y_t\leftarrow\arg\max_{a\in\mathcal{Y}}z_{t,a}$ \COMMENT{Eq.~(\ref{eq:decision_function})}
    \IF{$y_t=\texttt{uncertain}$ \AND $\bar d_t<d_{\mathrm{close}}$}
        \STATE $y_t\leftarrow\arg\max_{a\in\{\texttt{yes},\texttt{no}\}}z_{t,a}$ \COMMENT{close-range fallback}
    \ENDIF
    \IF{$y_t=\texttt{yes}$}
        \STATE Call the stop action; \textbf{return}
    \ELSIF{$y_t=\texttt{no}$}
        \STATE Reject the candidate and resume exploration; \textbf{return}
    \ENDIF
    \STATE If no waypoints remain, sample $\mathbf{A}_t\sim\pi_\phi(\cdot\mid M_t,D_t)$
    \STATE Execute one action toward the next waypoint; $t\leftarrow t+1$
\ENDWHILE
\STATE Resume exploration \COMMENT{track lost}
\end{algorithmic}
\end{algorithm}

%% file: sections/experiments.tex
\section{Experiments}
\label{sec:experiments}

We first report ObjectNav performance across systems, then analyze closed-loop verification and its components, and finally examine efficiency and remaining failures.

\subsection{Experimental Setup}
\label{subsec:exp_setup}

\textbf{Datasets and evaluation splits.}
We evaluate navigation in Habitat on the standard validation episodes of Gibson~\cite{xia2018gibson} (1{,}000 episodes), HM3D v1 (2{,}000), and HM3D v2~\cite{ramakrishnan2021hm3d,yadav2023hm3dsem} (1{,}000). The verification stack is trained, refined, validated, and tested on scene-disjoint HM3D splits, and the approach policy is trained on expert demonstrations from the same training scenes. MP3D~\cite{chang2017matterport3d} uses a 21-category protocol, whereas the stack is trained on the six categories shared by Gibson and HM3D, so MP3D serves only for the candidate-level distribution-shift evaluation of Table~\ref{tab:verifier_shift}.

\textbf{Metrics.}
At the episode level, we report success rate (SR) and Success weighted by Path Length (SPL)~\cite{anderson2018spl}; closed-loop results are means over three evaluation seeds. Every failed episode is attributed to exactly one cause, in priority order: false stop (FS), where the episode terminates by accepting a non-target; false reject (FR), where at least one true-target candidate is detected and then rejected or abandoned; exploration failure, where no true target is detected within the budget; or other, covering timeouts with a pending candidate and collision-induced termination. Episode-level FS and FR are therefore exclusive shares of all episodes. Candidate-level FS and FR are instead class-conditional rates: FS is the fraction of non-target candidate events accepted with \texttt{yes}, and FR is the fraction of target candidate events rejected with \texttt{no}; events that end in abandonment or episode termination with a pending candidate remain in the denominator but enter neither numerator. In the approach phase, a candidate is correctly resolved only if a correct \texttt{yes}/\texttt{no} decision is reached before the close-range fallback; fallback-triggered decisions, track loss, abandonment, and episode termination with a pending candidate do not count as correct resolution.

\textbf{Baselines and system integration.}
The eight baselines span end-to-end and modular ObjectNav systems based on learned policies, semantic exploration, frontier scoring, value maps, topological reasoning, and target-centric fusion. CLOSER attaches to each as a self-contained verification layer that runs its own Grounded-SAM, ByteTrack, and SAM (Sec.~\ref{subsec:closed_loop_inference}) on the RGB-D stream of the baseline, so every system receives candidates from the same detection-and-tracking pipeline. Exploration, mapping, and the low-level planner remain unchanged, and CLOSER replaces only the native stopping rule. The 500-step budget, action space, and success criterion of each dataset are likewise unchanged, so verification steps, including the approach phase, count against the same budget. Simulator annotations are used only for training labels and privileged expert construction. Systems with an LLM decision module use the same locally deployed \texttt{Qwen3-30B-A3B-Instruct-2507} model across datasets.

\textbf{Implementation details.}
The perceiver is trained on 78{,}350 labeled single-view detections from 108 HM3D training scenes (60{,}030 natural Grounded-SAM detections and 18{,}320 rendered samples covering the two rare regimes of Sec.~\ref{subsec:render_augmentation}), and the fusion MLP on 8{,}070 tracked multi-view sequences. Table~\ref{tab:implementation_parameters} lists the remaining settings; the view-diversity threshold $\delta_{\mathrm{view}}$ is selected on the validation split.

\begin{table}[t]
\centering
\caption{Implementation settings for CLOSER.}
\label{tab:implementation_parameters}
\begin{tabular}{@{}lr@{}}
\toprule
Setting & Value \\
\midrule
Target-overlap thresholds $(\eta_+,\eta_-)$ & $(0.25,\;0.05)$ \\
Minimum mask-to-image ratio $\alpha$ & $0.02$ \\
Minimum box-fill ratio $\beta$ & $0.40$ \\
Maximum mean mask depth $\gamma$ & $3.0$\,m \\
\addlinespace[2pt]
Cache capacity $K$ & $4$ views \\
View-diversity threshold $\delta_{\mathrm{view}}$ & $0.18$ \\
Cached-view order & Oldest to newest \\
Fusion MLP input & $4\times3$ raw probabilities \\
\addlinespace[2pt]
LoRA rank / scaling factor & $64 / 128$ \\
LoRA target modules & Attention and MLP projections \\
Perceiver SFT learning rate & $2\times10^{-5}$ \\
Refinement correct-to-misclassified ratio & $1.5$ \\
\addlinespace[2pt]
Expert B\'ezier degree & $7$ \\
Waypoint horizon $H$ & $8$ \\
Waypoint spacing & $0.25$\,m \\
DDIM denoising steps & $10$ \\
Policy learning rate & $1\times10^{-4}$ \\
\addlinespace[2pt]
Fallback mean-depth threshold $d_{\mathrm{close}}$ & $1.0$\,m \\
\bottomrule
\end{tabular}
\end{table}

\subsection{Overall ObjectNav Performance}
\label{subsec:navigation_results}

\begin{figure*}[t]
\centering
\includegraphics[width=\textwidth]{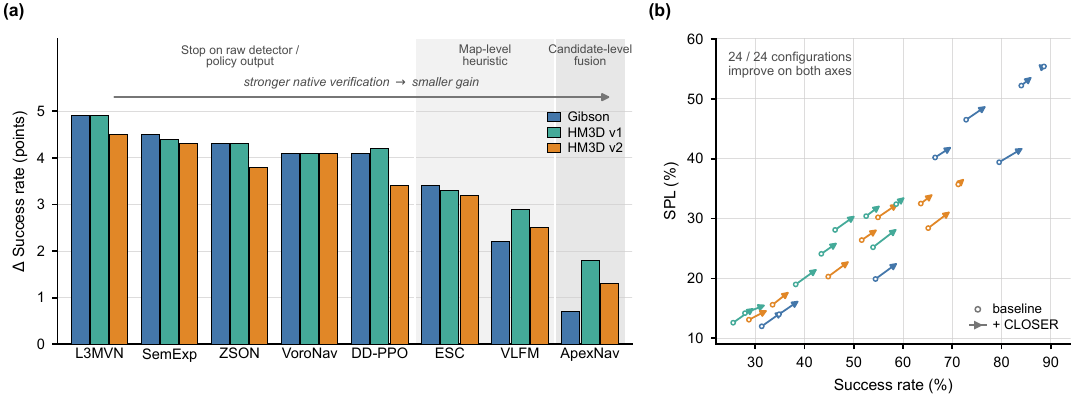}
\caption{\textbf{Paired ObjectNav gains from CLOSER.} (a) Absolute SR gains grouped by dataset and by the native candidate-verification mechanism (shading, left to right: none, map-level heuristic, and candidate-level fusion). (b) Changes in SR and SPL for the same 24 system--dataset pairs; each arrow starts at the baseline and ends at its ``+ CLOSER'' result. Data from Table~\ref{tab:navigation_harness_gains}.}
\label{fig:plugin_gains}
\end{figure*}

Table~\ref{tab:navigation_harness_gains} compares eight ObjectNav systems with their CLOSER variants on Gibson, HM3D v1, and HM3D v2; each ``+ CLOSER'' row differs from its baseline only in the stopping rule under the protocol of Sec.~\ref{subsec:exp_setup}. CLOSER improves both SR and SPL in all 24 system--dataset pairs (Fig.~\ref{fig:plugin_gains}), with SR gains averaging 3.6 points (up to 4.9) and SPL gains averaging 2.0 points (up to 3.1). Since approach steps count against the unchanged 500-step budget, the consistent SPL gains show that the path spent on additional views is more than repaid by the failures they prevent.

\begin{table*}[t]
\centering
\caption{\textbf{End-to-end ObjectNav results.} SR/SPL of eight ObjectNav systems and their CLOSER plug-in variants on Gibson, HM3D v1, and HM3D v2 (OV: open-vocabulary). Each shaded ``+ CLOSER'' row reports the absolute gain over its baseline in parentheses; CLOSER results are means over three seeds (standard deviations $\leq$0.9 SR and $\leq$0.6 SPL points). CLOSER improves both metrics for every system on every dataset.}
\label{tab:navigation_harness_gains}
\renewcommand{\arraystretch}{1.2}
\small
\begin{tabular*}{\textwidth}{@{}l@{\hspace{14pt}}c@{\extracolsep{\fill}}c@{\extracolsep{0pt}\hspace{16pt}}c@{\extracolsep{\fill}}c@{\extracolsep{0pt}\hspace{16pt}}c@{\extracolsep{\fill}}c@{\extracolsep{0pt}\hspace{16pt}}c@{}}
\toprule
\multirow{2}{*}{Method} & \multirow{2}{*}{OV} &
\multicolumn{2}{c}{Gibson} &
\multicolumn{2}{c}{HM3D v1} &
\multicolumn{2}{c}{HM3D v2} \\
\cmidrule(lr){3-4}\cmidrule(lr){5-6}\cmidrule(lr){7-8}
& & SR(\%) $\uparrow$ & SPL(\%) $\uparrow$ & SR(\%) $\uparrow$ & SPL(\%) $\uparrow$ & SR(\%) $\uparrow$ & SPL(\%) $\uparrow$ \\
\midrule
\multicolumn{8}{@{}l}{\textit{Plug-in comparison}} \\
ZSON~\cite{majumdar2022zson} & \checkmark & 31.3 & 12.0 & 25.5 & 12.6 & 28.7 & 13.1 \\
\closerband
\quad + CLOSER & \checkmark & \gaincell{35.6}{4.3} & \gaincell{14.5}{2.5} & \gaincell{29.8}{4.3} & \gaincell{15.1}{2.5} & \gaincell{32.5}{3.8} & \gaincell{14.7}{1.6} \\
\addlinespace[2pt]
DD-PPO~\cite{wijmans2020ddppo} & $\times$ & 34.8 & 14.0 & 27.9 & 14.2 & 33.5 & 15.6 \\
\closerband
\quad + CLOSER & $\times$ & \gaincell{38.9}{4.1} & \gaincell{16.3}{2.3} & \gaincell{32.1}{4.2} & \gaincell{15.6}{1.4} & \gaincell{36.9}{3.4} & \gaincell{17.8}{2.2} \\
\addlinespace[2pt]
ESC~\cite{zhou2023esc} & \checkmark & 66.5 & 40.2 & 43.4 & 24.1 & 51.6 & 26.4 \\
\closerband
\quad + CLOSER & \checkmark & \gaincell{69.9}{3.4} & \gaincell{42.0}{1.8} & \gaincell{46.7}{3.3} & \gaincell{26.0}{1.9} & \gaincell{54.8}{3.2} & \gaincell{28.2}{1.8} \\
\addlinespace[2pt]
VoroNav~\cite{wu2024voronav} & \checkmark & 72.8 & 46.5 & 46.2 & 28.1 & 54.9 & 30.2 \\
\closerband
\quad + CLOSER & \checkmark & \gaincell{76.9}{4.1} & \gaincell{48.8}{2.3} & \gaincell{50.3}{4.1} & \gaincell{30.5}{2.4} & \gaincell{59.0}{4.1} & \gaincell{32.5}{2.3} \\
\addlinespace[2pt]
VLFM~\cite{yokoyama2024vlfm} & \checkmark & 84.0 & 52.2 & 52.5 & 30.4 & 63.6 & 32.5 \\
\closerband
\quad + CLOSER & \checkmark & \gaincell{86.2}{2.2} & \gaincell{53.7}{1.5} & \gaincell{55.4}{2.9} & \gaincell{32.2}{1.8} & \gaincell{66.1}{2.5} & \gaincell{34.1}{1.6} \\
\addlinespace[2pt]
L3MVN~\cite{yu2023l3mvn} & $\times$ & 79.5 & 39.4 & 53.9 & 25.2 & 65.1 & 28.4 \\
\closerband
\quad + CLOSER & $\times$ & \gaincell{84.4}{4.9} & \gaincell{41.8}{2.4} & \gaincell{58.8}{4.9} & \gaincell{28.3}{3.1} & \gaincell{69.6}{4.5} & \gaincell{31.2}{2.8} \\
\addlinespace[2pt]
SemExp~\cite{chaplot2020semexp} & $\times$ & 54.4 & 19.9 & 38.2 & 19.0 & 44.8 & 20.3 \\
\closerband
\quad + CLOSER & $\times$ & \gaincell{58.9}{4.5} & \gaincell{22.6}{2.7} & \gaincell{42.6}{4.4} & \gaincell{21.6}{2.6} & \gaincell{49.1}{4.3} & \gaincell{22.9}{2.6} \\
\addlinespace[2pt]
ApexNav~\cite{zhang2025apexnav} & \checkmark & 88.6 & 55.4 & 58.6 & 32.4 & 71.2 & 35.7 \\
\closerband
\quad + CLOSER & \checkmark & \gaincell{89.3}{0.7} & \gaincell{55.6}{0.2} & \gaincell{60.4}{1.8} & \gaincell{33.6}{1.2} & \gaincell{72.5}{1.3} & \gaincell{36.7}{1.0} \\
\bottomrule
\end{tabular*}
\end{table*}

The size of each gain reflects how much candidate verification the baseline already performs. DD-PPO, ZSON, SemExp, L3MVN, and VoroNav stop on detector or policy outputs and gain 3.4--4.9 SR points on each dataset; ESC and VLFM use map-level scoring without a candidate-level check and gain 2.2--3.4 points; ApexNav already fuses passively acquired views of the candidate~\cite{zhang2025apexnav} and gains the least, yet CLOSER still improves its SR and SPL on all three datasets and makes it the strongest system in Table~\ref{tab:navigation_harness_gains}. This ordering is consistent with attributing the baseline failures to the stopping decision (Sec.~\ref{subsec:failure_cause_analysis}), and the gains over ApexNav indicate that the views gathered actively during the approach phase add evidence that passive acquisition does not provide.

\subsection{Failure Cause Analysis}
\label{subsec:failure_cause_analysis}

We analyze native VLFM on HM3D v2 using the episode-level failure attribution of Sec.~\ref{subsec:exp_setup}. Its SR is 63.6\%, with 19.4\% false stops and 13.8\% false rejects. These stop-attributable causes sum to 33.2 percentage points of the overall 36.4\% failure rate, so the stopping decision accounts for over 90\% of the failures; exploration and other failures account for the remaining 3.2 points.

CLOSER reduces the stop-attributable failures by more than half, from 33.2 to 15.7 points, with false stops falling to 6.4\% and false rejects to 9.3\%, and SR increases to 66.1\%. Preventing an early false stop or false reject allows navigation to continue, and some of the continued episodes subsequently time out, collide, or end with a pending candidate. The exploration/other share therefore rises from 3.2\% to 18.2\%, and more than half of the remaining failures fall outside the stopping decision that CLOSER addresses.

\subsection{Qualitative Results}
\label{subsec:qualitative_results}

\begin{figure}[t]
\centering
\includegraphics[width=\columnwidth]{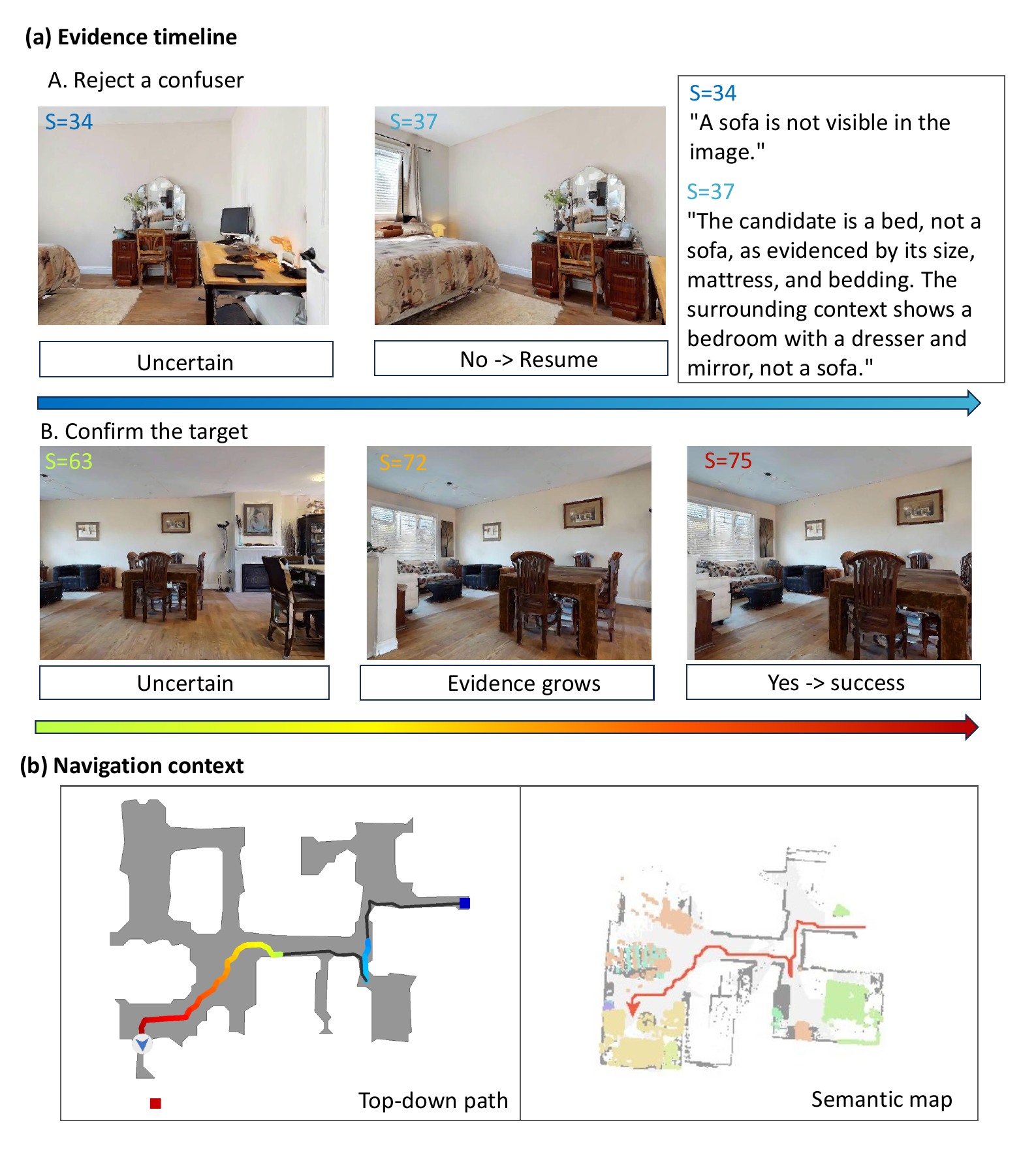}
\caption{\textbf{Closed-loop verification in one successful episode with target category \emph{sofa}.} (a) Evidence timeline of the two candidate events (S: step index). In event A, the initial view is deferred as \texttt{uncertain}, and the view retained at step 37 reveals the candidate to be a bed, so the agent discards the candidate and resumes exploration. In event B, the fused decision remains \texttt{uncertain} at step 72 and becomes \texttt{yes} at step 75, so the agent stops. The quoted text shows the perceiver rationales for the two views of event A, decoded here for illustration. (b) Top-down path and semantic map of the episode; the colored path segments correspond to the timeline arrows in (a).}
\label{fig:qualitative}
\end{figure}

Fig.~\ref{fig:qualitative} shows how the closed loop prevents both error types in one successful episode with target category \emph{sofa}. In the first candidate event, the fused decision on the initial view at step 34 is \texttt{uncertain}, so the agent enters the approach phase. The view retained at step 37 shows that the candidate is a bed; the fused decision becomes \texttt{no}, and the agent discards the candidate and resumes exploration instead of ending the episode with a false stop. In the second event, the initial view at step 63 is likewise deferred, and two further views retained at steps 72 and 75 add evidence during the approach phase until the fused decision becomes \texttt{yes}. The agent then calls the stop action, and the episode succeeds. In the navigation context of Fig.~\ref{fig:qualitative}(b), the colored segments of the top-down path mark the steps spanned by the two events, and the black segments correspond to exploration by the baseline. Neither map in the panel is an input to CLOSER (Sec.~\ref{subsec:active_sensing}).

\begin{figure*}[!t]
\centering
\includegraphics[width=0.9\textwidth]{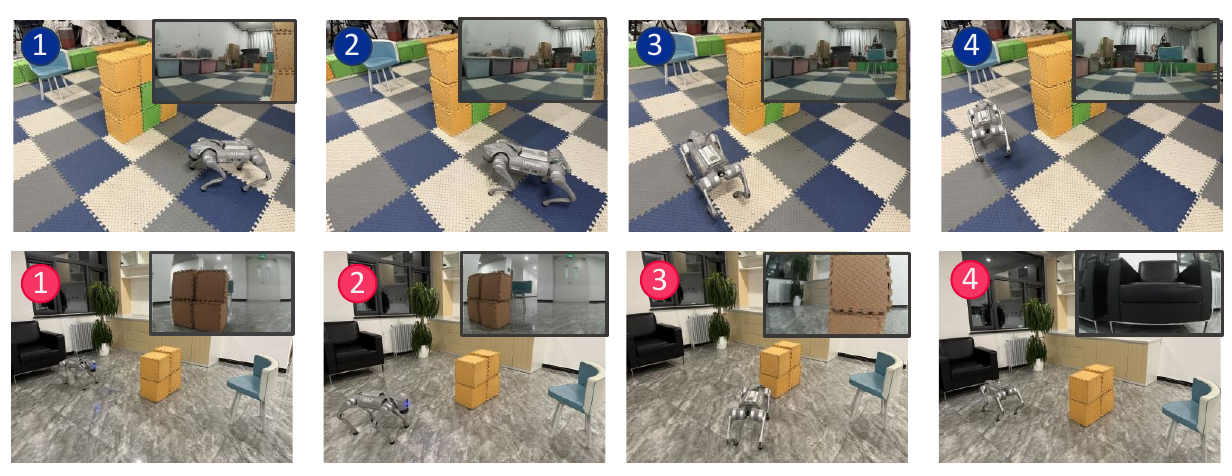}
\caption{\textbf{Real-world verification on a quadruped robot.} The verification stack runs on the onboard camera stream of a Unitree Go2 in two indoor scenes (one scene per row, frames ordered left to right; insets show the onboard view). Stacked blocks occlude the candidate (a chair, top; an armchair, bottom), so the initial views are deferred as \texttt{uncertain}. As the robot moves around the blocks, clearer views enter the cache, and the fused decision becomes decisive once an unobstructed view is retained (rightmost insets).}
\label{fig:realworld}
\vspace{-3mm}
\end{figure*}

\textbf{Real-world verification.}
Fig.~\ref{fig:realworld} shows the verification stack deployed on a Unitree Go2 quadruped in two real indoor scenes, running the same components as in Sec.~\ref{subsec:closed_loop_inference} on the onboard camera stream without any fine-tuning on real data. In both scenes, stacked blocks occlude the candidate at the initial viewpoint, so the initial views are deferred; as the robot moves around the blocks, clearer views enter the cache, and the fused decision becomes decisive once the candidate is fully visible. The tri-state verification thus transfers from simulation to real camera streams without adaptation.

\subsection{Ablation Studies}
\label{subsec:ablation_studies}

\subsubsection{Closed-loop Verification}

Table~\ref{tab:closed_loop_ablation} compares the native VLFM stopping rule, two binary verifiers, tri-state verification without the approach policy, and the complete system on the same HM3D v2 episodes. Binary GPT-4o-mini follows the API-based verification of prior work~\cite{zhu2025strive}; the local binary variant uses the same cache implementation but has no \texttt{uncertain} state, so its first decision terminates the candidate event. The static tri-state variant keeps the candidate tracked while VLFM continues to explore and recomputes the fused decision whenever the cache admits a new view, without a dedicated approach policy; the close-range fallback still applies if VLFM brings the agent within 1\,m of the candidate. Cache construction and approach policies are evaluated separately in Tables~\ref{tab:keyframe_ablation} and~\ref{tab:active_policy_comparison}.

\begin{table}[t]
\centering
\caption{\textbf{Closed-loop ablation of the verification layer.} All variants are built on VLFM and evaluated on the HM3D v2 validation episodes. FS / FR / FB: episode-level false stop, false reject, and fallback trigger (\%).}
\label{tab:closed_loop_ablation}
\setlength{\tabcolsep}{3.5pt}
\begin{tabular}{@{}lccccc@{}}
\toprule
Variant & SR$\uparrow$ & SPL$\uparrow$ & FS$\downarrow$ & FR$\downarrow$ & FB$\downarrow$ \\
\midrule
VLFM native stop & 63.6 & 32.5 & 19.4 & 13.8 & -- \\
Binary GPT-4o-mini (API) & 64.9 & 33.3 & 11.2 & 12.1 & -- \\
Binary local & 64.4 & 33.0 & 13.6 & 14.2 & -- \\
Tri-state + diverse cache (static) & 65.3 & 33.6 & 8.7 & 15.1 & 4.2 \\
\textbf{+ Diffusion policy (CLOSER)} & 66.1 & 34.1 & 6.4 & 9.3 & 9.8 \\
\bottomrule
\end{tabular}
\end{table}

\begin{figure}[t]
\centering
\includegraphics[width=\columnwidth]{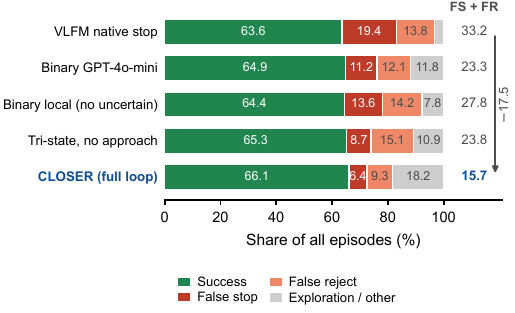}
\caption{\textbf{Episode outcomes of the closed-loop ablation.} Each episode is assigned to exactly one outcome; the right column shows the total stop-attributable error (FS $+$ FR). Data from Table~\ref{tab:closed_loop_ablation}.}
\label{fig:closed_loop}
\end{figure}

The binary API and local verifiers reduce false stops from 19.4\% to 11.2\% and 13.6\%, respectively, but improve SR by at most 1.3 points. The local binary variant also increases false rejects from 13.8\% to 14.2\%, since it has no deferral state and must assign a binary outcome when the available evidence is insufficient.

Adding an \texttt{uncertain} state without the approach policy further reduces false stops to 8.7\%, while false rejects increase to 15.1\%: deferred target candidates are abandoned when the views that VLFM subsequently observes do not provide sufficient evidence. With the approach policy, false stops and false rejects decrease to 6.4\% and 9.3\%, and SR/SPL reach 66.1/34.1 (Fig.~\ref{fig:closed_loop}). During the approach phase, the agent acquires additional views of deferred candidates, which accounts for the reduction in false rejects relative to the static tri-state variant. The complete system triggers the close-range fallback in 9.8\% of episodes; fallback-induced false stops and false rejects are included in the reported rates.

\subsubsection{Evidence-aware Verification and View Selection}
\label{subsec:standalone_verifier_benchmark}

Table~\ref{tab:standalone_verification} compares verifier variants on 5{,}184 held-out Grounded-SAM candidate events from the scene-disjoint HM3D test split, separating semantic decisions on clear observations from deferral decisions on insufficient ones. Binary SFT shares the backbone and corpus of the tri-state models but assigns each insufficient observation its latent semantic label~\cite{zhu2025strive}; the threshold variant defers when its maximum label-token probability falls below a threshold selected on the validation split.

\begin{table}[t]
\centering
\caption{\textbf{Verifier comparison on 5{,}184 held-out Grounded-SAM detections.} Clear columns: correct semantic decisions on clear observations; Low-Q columns: correct deferral of insufficient observations (zero for the pure binary verifier, which cannot defer). The lower block reports the CLOSER training stages, the first two on single views; parenthesized values give the gain over the previous stage. All values in \%.}
\label{tab:standalone_verification}
\setlength{\tabcolsep}{3pt}
\resizebox{\columnwidth}{!}{%
\begin{tabular}{@{}lcccc@{}}
\toprule
Verifier / variant &
\shortstack{Clear$\rightarrow$\\yes $\uparrow$} &
\shortstack{Low-Q tgt$\rightarrow$\\unc.\ $\uparrow$} &
\shortstack{Clear$\rightarrow$\\no $\uparrow$} &
\shortstack{Low-Q cnf$\rightarrow$\\unc.\ $\uparrow$} \\
\midrule
Qwen2.5-VL zero-shot & 68.7 & 42.3 & 55.1 & 39.8 \\
Binary SFT (\texttt{yes}/\texttt{no}) & 87.4 & 0.0 & 78.6 & 0.0 \\
Binary SFT + threshold & 79.8 & 82.1 & 69.4 & 80.6 \\
\midrule
Direct-label SFT, single view & 83.6 & 67.4 & 73.2 & 65.8 \\
+ Rendered regimes + rationale & \gaincell{85.9}{2.3} & \gaincell{78.7}{11.3} & \gaincell{77.1}{3.9} & \gaincell{77.3}{11.5} \\
\textbf{Full CLOSER stack} & \gaincell{89.1}{3.2} & \gaincell{85.4}{6.7} & \gaincell{84.6}{7.5} & \gaincell{83.8}{6.5} \\
\bottomrule
\end{tabular}%
}
\end{table}

Binary SFT reaches 87.4\% and 78.6\% on clear target and confuser observations but cannot defer; threshold deferral raises the two low-quality rates to 82.1\% and 80.6\% at the cost of the clear-observation rates, which drop to 79.8\% and 69.4\%. Among the CLOSER stages, rendered regimes with rationale supervision raise the average low-quality deferral of direct-label SFT from 66.6\% to 78.0\%, and sequence-level fusion with hard refinement brings the full stack to 85.4\% and 83.8\% while also improving the clear-observation rates. At the deployment operating point of Table~\ref{tab:runtime_cost}, the local stack reaches 86.1\% tri-state accuracy, compared with 82.4\% for the single-view GPT-4o-mini API comparator.

\textbf{Distribution shift.}
Table~\ref{tab:verifier_shift} evaluates threshold deferral and tri-state verification across three domains, with the binary threshold selected once on the HM3D validation split and kept fixed.

\begin{table}[t]
\centering
\caption{\textbf{Deferral quality under distribution shift.} The deferral threshold of the binary verifier is selected once on the HM3D validation split and kept fixed. Defer: fraction of insufficient-evidence candidates correctly deferred; FS / FR: candidate-level false stop and false reject. All values in \%.}
\label{tab:verifier_shift}
\setlength{\tabcolsep}{3pt}
\resizebox{\columnwidth}{!}{%
\begin{tabular}{lcccccc}
\toprule
\multirow{2}{*}{Evaluation domain} &
\multicolumn{3}{c}{Binary SFT + threshold} &
\multicolumn{3}{c}{CLOSER (tri-state)} \\
\cmidrule(lr){2-4}\cmidrule(lr){5-7}
& Defer $\uparrow$ & FS $\downarrow$ & FR $\downarrow$ & Defer $\uparrow$ & FS $\downarrow$ & FR $\downarrow$ \\
\midrule
HM3D held-out (in-domain) & 81.4 & 12.6 & 11.2 & 84.6 & 7.5 & 8.4 \\
MP3D candidates (cross-dataset) & 63.7 & 17.9 & 15.8 & 82.1 & 8.6 & 9.3 \\
Rendered hard regimes & 46.8 & 23.4 & 20.7 & 79.2 & 9.8 & 10.9 \\
\bottomrule
\end{tabular}%
}
\end{table}

\begin{figure}[t]
\centering
\includegraphics[width=\columnwidth]{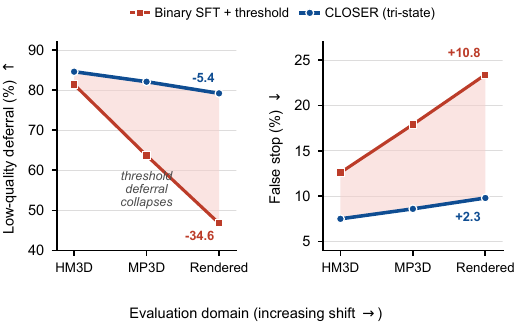}
\caption{\textbf{Deferral under distribution shift.} The fixed binary threshold and CLOSER are evaluated on held-out HM3D candidates, MP3D candidates, and the rendered hard regimes of Table~\ref{tab:two_axis}. Annotations show changes from the in-domain to the rendered setting; the shaded region shows the gap between the methods. Data from Table~\ref{tab:verifier_shift}.}
\label{fig:verifier_shift}
\end{figure}

From HM3D to MP3D and the rendered regimes, the deferral rate of the thresholded model drops from 81.4\% to 63.7\% and 46.8\%, whereas that of CLOSER drops only from 84.6\% to 82.1\% and 79.2\% (Fig.~\ref{fig:verifier_shift}); over the same range, the false-stop rate rises from 12.6\% to 23.4\% for the threshold but only from 7.5\% to 9.8\% for CLOSER. The explicitly supervised \texttt{uncertain} state is therefore far less sensitive to domain shift than a fixed confidence threshold.

\textbf{View selection.}
Table~\ref{tab:keyframe_ablation} holds the learned components fixed and compares a single latest frame with the latest $K$ frames and CLIP-diverse selection, both with $K{=}4$.

\begin{table}[t]
\centering
\caption{\textbf{Ablation of cache construction.} The learned components are fixed; multi-view rows use $K{=}4$. Acc.: tri-state accuracy; Low-Q: fraction of insufficient-evidence candidates correctly deferred; FS: candidate-level false stop. All values in \%.}
\label{tab:keyframe_ablation}
\setlength{\tabcolsep}{3.5pt}
\begin{tabular}{@{}lccc@{}}
\toprule
Evidence input & Acc.\ $\uparrow$ & Low-Q $\uparrow$ & FS $\downarrow$ \\
\midrule
Latest frame only & 80.7 & 78.9 & 11.4 \\
Latest-$K$ frames & 83.5 & 81.2 & 9.2 \\
\textbf{CLIP-diverse (ours)} & 86.1 & 84.6 & 7.5 \\
\bottomrule
\end{tabular}
\end{table}

The latest-$K$ cache raises tri-state accuracy from 80.7\% to 83.5\%, and CLIP-diverse selection raises it further to 86.1\% while lifting low-quality deferral to 84.6\% and lowering candidate-level false stops from 9.2\% to 7.5\%. Selecting complementary views thus improves the evidence supplied to the fusion MLP at the same cache budget.

\subsubsection{Approach Policy}
\label{subsec:active_sensing_ablation}

Table~\ref{tab:active_policy_comparison} compares policies on candidates initially assigned \texttt{uncertain}. All variants share the verifier, evidence rule, and close-range fallback and differ only in how they acquire additional observations; the fallback rate is conditioned on this subset and differs from the episode-level rate in Table~\ref{tab:closed_loop_ablation}.

\begin{table}[t]
\centering
\caption{\textbf{Approach-policy comparison on initially uncertain candidates.} Correct: correct \texttt{yes}/\texttt{no} decision before the close-range fallback; Confuser: rejection rate on initially uncertain confusers; FB: fallback trigger rate on this candidate subset. All values except Path are in \%.}
\label{tab:active_policy_comparison}
\setlength{\tabcolsep}{3pt}
\resizebox{\columnwidth}{!}{%
\begin{tabular}{@{}lcccc@{}}
\toprule
Policy & Correct $\uparrow$ & Confuser $\uparrow$ & Path (m) $\downarrow$ & FB $\downarrow$ \\
\midrule
In-place look-around & 32.4 & 26.1 & 0.00 & 0.0 \\
Straight-line policy & 61.7 & 54.3 & 2.84 & 37.6 \\
Viewpoint re-verif.\ (STRIVE)~\cite{zhu2025strive} & 70.2 & 63.8 & 2.21 & 25.4 \\
Voronoi--B\'ezier det.\ BC & 76.5 & 70.9 & 2.08 & 18.1 \\
\textbf{Voronoi--B\'ezier diffusion} & 83.6 & 78.4 & 1.69 & 12.7 \\
\bottomrule
\end{tabular}%
}
\end{table}

\begin{figure}[t]
\centering
\includegraphics[width=\columnwidth]{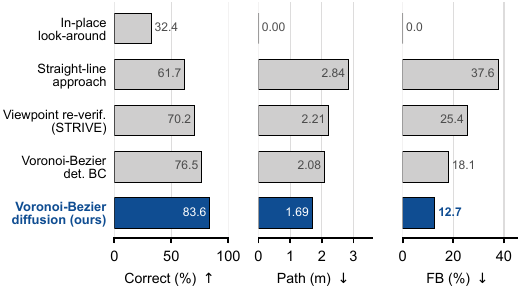}
\caption{\textbf{Resolution accuracy, extra path, and fallback rate of approach policies} on the initially \texttt{uncertain} candidate subset. Data from Table~\ref{tab:active_policy_comparison}.}
\label{fig:policy_pareto}
\vspace{-3mm}
\end{figure}

In-place look-around resolves only 32.4\% of candidates, and the straight-line policy raises correct resolution to 61.7\% at the cost of 2.84\,m of extra path and a 37.6\% fallback rate; selecting a single re-verification viewpoint as in STRIVE~\cite{zhu2025strive} reaches 70.2\% with 2.21\,m. The deterministic and diffusion policies imitate the same Voronoi--B\'ezier demonstrations, and the diffusion policy raises correct resolution from 76.5\% to 83.6\%, shortens the extra path from 2.08\,m to 1.69\,m, and lowers the fallback rate from 18.1\% to 12.7\% (Fig.~\ref{fig:policy_pareto}). It also rejects 78.4\% of initially uncertain confusers, although it is trained only on target-present demonstrations.

\subsection{Efficiency and Training Cost}
\label{subsec:efficiency}

Table~\ref{tab:runtime_cost} compares the single-view API and local perceivers with the complete verification stack. Local measurements use batch size~1 on one NVIDIA A100 40\,GB GPU and include visual preprocessing and decoding; API latency is measured end to end. Because the perceiver is queried once per retained view and the MLP fuses the cached probabilities after each cache update, the latency of the stack is reported per retained view.

\begin{table}[t]
\centering
\caption{\textbf{Verifier runtime and deployment cost.} Local VLM latency and memory are measured on a single NVIDIA A100 40\,GB GPU; API cost is in USD per 1{,}000 queries. Latency of the perceiver--MLP stack is per retained view; each MLP fusion adds 0.1\,ms.}
\label{tab:runtime_cost}
\setlength{\tabcolsep}{3pt}
\resizebox{\columnwidth}{!}{%
\begin{tabular}{lcccccc}
\toprule
Verifier & Local & Views & \shortstack{Tri-state\\accuracy $\uparrow$} & \shortstack{Latency\\(s) $\downarrow$} & \shortstack{Memory\\(GB) $\downarrow$} & \shortstack{Cost/1k\\queries $\downarrow$} \\
\midrule
GPT-4o-mini API (tri-state) & No & 1 & 82.4 & 1.38 & -- & 0.41 \\
Qwen2.5-VL zero-shot & Yes & 1 & 52.6 & 0.31 & 16.9 & 0 \\
CLOSER single-view perceiver & Yes & 1 & 80.7 & 0.28 & 17.2 & 0 \\
\textbf{CLOSER perceiver + raw-probability MLP} & Yes & $\leq4$ & 86.1 & 0.28/view & 17.2 & 0 \\
\bottomrule
\end{tabular}%
}
\end{table}

The single-view perceiver requires 0.28\,s and 17.2\,GB per retained view, compared with 1.38\,s and \$0.41 per 1{,}000 queries for GPT-4o-mini. Fusing up to four cached views adds 0.1\,ms and raises tri-state accuracy from 80.7\% to 86.1\%, above the 82.4\% of the API comparator, while leaving per-view latency and peak memory unchanged, since each retained view is evaluated exactly once.

\textbf{Training cost.}
LoRA adaptation updates 160.8\,M parameters (2.1\% of the backbone). The perceiver and policy stages require 111 GPU-hours on four A100-80GB GPUs: 76 hours for first-round SFT on the 78{,}350 single-view samples, 19 hours for hard-example refinement, and 16 hours for diffusion-policy training on 14{,}680 demonstration chunks (fusion-MLP training excluded). The triage rule of Eq.~(\ref{eq:cot_triage}) queries the GPT-4o teacher for 42.3\% of the corpus and reduces annotation cost by $2.4\times$ (\$518 vs.\ \$1{,}226).

\subsection{Remaining Failure Analysis}
\label{subsec:failure_analysis}

Table~\ref{tab:confuser_pairs} compares the conditional candidate-level false-stop rates of the detector-only rule, the binary SFT verifier, and CLOSER on four non-target confusion subsets.

\begin{table}[t]
\centering
\caption{\textbf{Candidate-level false-stop rate (\%) on characteristic confuser pairs} (VLFM+CLOSER, HM3D v2). ``A as B'' denotes a candidate of category A detected while searching for target category B; ``Other confusions'' pools the unlisted non-target confusion types.}
\label{tab:confuser_pairs}
\setlength{\tabcolsep}{4pt}
\begin{tabular}{@{}lccc@{}}
\toprule
Confusion & \shortstack{Detector\\only} & \shortstack{Binary\\verifier} & CLOSER \\
\midrule
Sofa as bed & 79.2 & 53.1 & 35.4 \\
Stool as chair & 70.6 & 45.3 & 28.8 \\
Sink as toilet & 64.8 & 42.7 & 27.1 \\
Other confusions & 46.3 & 30.2 & 19.6 \\
\bottomrule
\end{tabular}
\end{table}

For the sofa--bed, stool--chair, and sink--toilet confusions, binary verification reduces false stops from 64.8--79.2\% to 42.7--53.1\%, and CLOSER further reduces them to 27.1--35.4\%; on the remaining confusions, the rates are 46.3\%, 30.2\%, and 19.6\%. Binary verification handles clear confusers, while the closed loop acquires additional views for candidates that remain uncertain. As each row is conditioned on a different subset, these rates are not comparable with the aggregate 7.5\% false-stop rate of Tables~\ref{tab:verifier_shift} and~\ref{tab:keyframe_ablation}.

On VLFM with CLOSER, the remaining episode-level false stops and false rejects are 6.4\% and 9.3\%, respectively (Table~\ref{tab:closed_loop_ablation}). Qualitative inspection identifies three recurrent modes.

\emph{Track switches between adjacent similar instances.} ByteTrack can transfer a candidate identity to a neighboring object with a similar appearance. The evidence cache may then mix views of different instances.

\emph{Persistent occlusion at close range.} When a candidate remains occluded within the 1\,m region, the binary fallback makes a decision from insufficient evidence. The episode-level fallback trigger rate is 9.8\%, while 19.7\% of fallback-triggered decisions on initially uncertain candidates are incorrect; these rates use different denominators.

\emph{Confusers with similar appearance and context.} Some instances, such as daybeds staged as sofas, remain ambiguous at close range and require cues beyond local appearance.

The remaining 18.2\% of episodes belong to the combined exploration/other category. They include episodes in which no target is detected within the budget, timeouts with a pending candidate, and collision-induced termination.

%% file: sections/conclusion.tex
\section{Conclusion}
\label{sec:conclusion}

We presented CLOSER, an evidence-aware active visual verification framework that treats the terminal decision of object-goal navigation as a closed-loop evidence acquisition problem rather than a one-shot binary filter. A CLIP gate maintains four complementary views, a locally deployable single-view VLM reports their raw tri-state probabilities, and a lightweight MLP learns \emph{when} the chronologically ordered cache supports a category commitment; a privileged-to-observation Diffusion Transformer learns \emph{how} to acquire a more informative view from only the tracked mask and depth. Inserted as a plug-in layer, CLOSER consistently improves success rate and path efficiency across heterogeneous ObjectNav harnesses; ablations show that the full gain requires the closed loop, since deferral without resolution reduces false stops but raises false rejects. Looking ahead, extending evidence-aware deferral from candidate verification to exploration commitments, and validating the mask-and-depth policy on physical platforms, are natural next steps toward irreversible decisions under partial observability.